# HaptiStylus: A Novel Stylus Capable of Displaying Movement and Rotational Torque Effects

Atakan Arasan, Cagatay Basdogan, T. Metin Sezgin

aarasan@ku.edu.tr, cbasdogan@ku.edu.tr, mtsezgin@ku.edu.tr

Koc University, Istanbul, Turkey, 34450

**Abstract -** With the emergence of pen-enabled tablets and mobile devices, stylus-based interaction has been receiving increasing attention. Unfortunately, styluses available in the market today are all passive instruments that are primarily used for writing and pointing. In this paper, we describe a novel stylus capable of displaying certain vibro-tactile and inertial haptic effects to the user. Our stylus is equipped with two vibration actuators at the ends, which are used to create a tactile sensation of up and down movement along the stylus. The stylus is also embedded with a DC motor, which is used to create a sense of bidirectional rotational torque about the long axis of the pen. Through two psychophysical experiments, we show that, when driven with carefully selected timing and actuation patterns, our haptic stylus can convey movement and rotational torque information to the user. Results from a further psychophysical experiment provide insight on how the shape of the actuation patterns effects the perception of rotational torque effect. Finally, experimental results from our interactive pen-based game show that our haptic stylus is effective in practical settings.

# 1. INTRODUCTION

There is a large and growing set of mobile applications, and in particular games, that use stylus-based input for manipulating and interacting with virtual objects. Despite the popularity of stylus as an input device, stylus-based interaction has not reached its potential, partly because styluses available in the market are passive instruments. These instruments were primarily designed for writing and pointing on mobile devices, and they lack the ability to create supportive feedback during interaction. To this end, we designed a novel stylus that is capable of creating two tactile effects in an effort to enrich stylus-based interaction. In particular, we built an active stylus that is capable of displaying movement and rotational torque effects.

Our design, which we named HaptiStylus, is equipped with two vibration motors positioned at the ends of its casing. Asynchronous actuation of these motors results in a sensory illusion known as "apparent tactile motion" [1]. With appropriate combination of stimulus duration and inter-stimulus onset interval (ISOI) of the vibration actuators, we show that a movement effect can be displayed along the body of the stylus.

The other effect that HaptiStylus is capable of displaying is rotational torque effect about the long axis of the stylus. A high torque rated DC motor, positioned underneath the fingers enables us to create a sense of clockwise and counter clockwise rotation. Powering up the DC motor creates a torque on the casing of the stylus. When the voltage pulse is cut off, the motor displays a reaction torque in the opposite direction. In our previous work, we showed that the torque created during start-up dominates over the reaction torque, and creates a sense of rotation in the intended direction [2]. Here, we further investigate the effects of the input waveform patterns on the torque created by the motor and the perceived sense of rotation.

In this paper, we report detailed results from three psychophysical experiments that explore the effects of actuation timing and waveform on the perception of movement and rotational torque effects. In particular, 1) we explore the effects of duration of actuation, and ISOI on the movement effect, and 2) the onset/offset durations and input waveform patterns on the rotational torque effect. In addition, we report results from our assessment of our haptic stylus in a game, designed to take advantage of the proposed haptic effects. This assessment is carried out by a more compact version of our stylus (HaptiStylus 2.0), which is fitted with a Bluetooth module for communication and an active stylus tip to enable fine grained tracking on a digitizing tablet. The results of three psychophysical experiments show that, with carefully selected parameters, we can create highly perceptible movement and rotational torque effects. We do so not only in a targeted perception experiment, but also in an immersive game where users' primary focus is not on perceiving effects, but on the game dynamics.

## 2. RELATED WORK

There are a few studies in the literature about displaying haptic effects on styluses through tactile channel. Lee et al. developed the first haptic stylus that displays forces along its longitudinal axis by means of a solenoid actuator at the tip [3]. Kyung and Jun-Seok designed

Ubi-Pen which displays vibration cues to the user via a single vibration motor and additionally, texture information at the fingertip through an embedded pin array [4]. Later, Kyung et al. miniaturized this pen by removing the embedded pin array [5]. They also developed a GUI on a touch screen to display vibrotactile haptic feedback to a user via a single linear actuator for user events such as button click, icon/file pickup, drag-and-drop, window resizing and scroll, text highlighting, and menu movement/selection. Kamuro et al. developed an ungrounded stylus to display force feedback to a user without the use of mechanical linkages [6]. Wintergerst et al. proposed a stylus, with a magnetically operated brake at the tip, to display haptic effects by controlling the friction between the tip and a display surface [7]. Withana et al. developed a haptic stylus that dynamically changes its effective length to create an illusion of penetrating into a display surface [8]. Poupyrev et al. explored the benefits of using vibrotactile haptic feedback displayed through a touch screen in pen-based interactions [9]. They placed four piezo actuators at the corners of pen-based touch display in between the LCD and the protective glass panel and displayed vibrotactile haptic feedback to a user in varying amplitude and frequency. The results of the user study showed that the subjects preferred haptic feedback when it was combined with an active gesture as in dragging a slider or highlighting a text using a pen interface. Our work complements this body of work by the virtue of focusing on different set of haptic effects (movement and rotational effects).

The haptic movement effect, tactile illusion of creating a continuous motion, was investigated first by Sherrick and Rogers via two vibration motors separated by a distance and placed on the skin of thigh [1]. They adjusted the stimulus duration and the delay between the actuation times to create an effect of travelling haptic stimulus. Tan and Pentland developed a wearable tactile chair equipped with vibration motors to display static lines and two-dimensional geometric patterns to the back of a user [10]. Israr and Poupyrev proposed an algorithm to display two-dimensional dynamic haptic effects to the back of a user sitting on a chair through a matrix of vibration actuators [11]. In another line of work, Israr and Poupyrev investigated the control parameter space for producing reliable continuous moving patterns on the forearm and back via haptic movement effect [12]. The results of the study showed that the ISOI space for forearm was influenced by both the motion direction and spacing of the actuators, while the ISOI space for back was only affected by the direction of actuation. Kim et al. created a sensation of travelling wave between two vibration actuators embedded in a cell phone by adjusting the magnitude and timing of the actuators [13]. Lim et al. used frequency modulation to display haptic movement effect between two hands [14]. Our work explores the use of haptic movement effect (apparent tactile motion illusion) on a pen-based interface that can be used with a tablet or a mobile device. In our stylus, the actuators do not come in direct contact with the skin, and the vibrations are transmitted indirectly through the stylus casing.

Compared to the haptic movement effect, the haptic rotational torque effect has been investigated less in the literature. Winfree et al. developed iTorqU, an ungrounded handheld device, to create directional torque feedback by means of a flywheel inside of a two-axis actuated gimbal [15]. Porquis et al. created a torque illusion via suction by using a vacuum driven interface embedded in a stylus [16]. Using a rotating flywheel, Amemiya and Gomi displayed directional torque feedback through a mobile device [17]. Unlike these studies, we simply utilize consecutive torque pulses displayed through a DC motor attached to the casing of our pen-based stylus to generate a perceptual sense of bidirectional rotation.

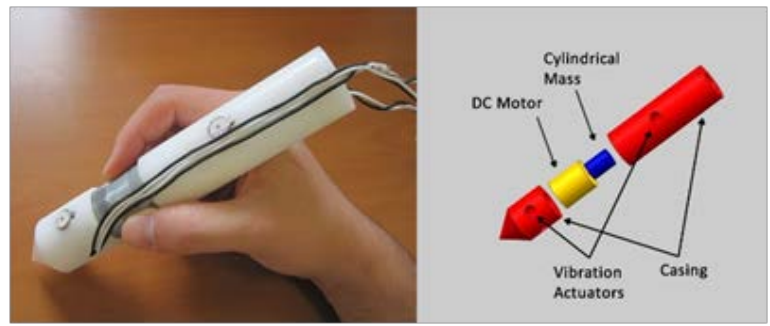

a)

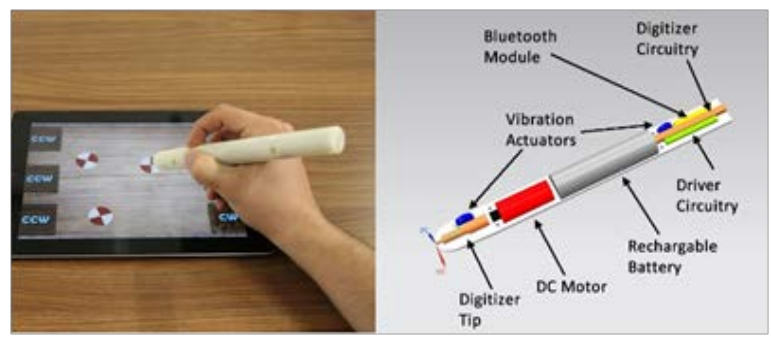

b)

Figure 1 – Physical components of HaptiStylus 1.0 (above) and HaptiStylus 2.0 (below)

To our knowledge, the application of the proposed haptic effects (haptic movement and rotation) on a pen-based interface is new. While the concept of haptic movement effect (i.e. apparent tactile motion) is not new and has been applied to the other parts of body such as forearm and back as discussed above, the haptic rotation effect is completely new. Moreover, we should emphasize that the earlier pen-based interfaces with haptic feedback has focused on displaying haptic effects for 2D GUI events such as button click, drag and drop, menu selection, etc. However, the proposed stylus and the haptic effects open new avenues for 3D haptic interaction (as discussed in the "Application" section) since the proposed stylus includes multiple actuators along its body.

3. METHOD

3.1 Design of the HaptiStylus

Our haptic stylus consists of a plastic cylindrical casing and three physical actuators: one DC motor for creating a sense of rotation, and two vibration actuators for generating movement effect. Two physical embodiments of our design are shown in Figure 1. The pen in Figure 1.a was designed as a low-cost embodiment, which doesn't include any circuitry inside, whereas the embodiment in Figure 1.b was designed as a standalone device with an embedded digitizing stylus tip, control circuitry for actuators, and a Bluetooth module for wireless communication (Table 1). The first design was used to evaluate the effectiveness of our approach for generating movement and rotational torque effects, which was further verified with the second design in a more practical computer game setting. The second design was also used to further investigate how the perception of rotational torque effect is influenced by the shape of the waveform that drives the actuators.

We utilize two coin-type eccentric rotating mass (ERM) actuators to display a movement effect through the stylus. The coin type ERM actuators are compact, lightweight and have a larger frequency and amplitude bandwidth than some other vibration actuators. Two coin type ERM actuators are placed close to the ends of the styluses (Figure 1). We used specialized haptic drivers to improve the performance of the vibration actuators. The drivers offer shorter start-up duration (0.1 ms) using only a single-ended pulse width modulation (PWM) input signal. The use of specialized driver circuits proved to be critical for achieving the targeted effects.

In order to display the rotational torque effect, we used a high-torque rated DC motor. The DC motor is positioned to coincide with the position of the fingers while holding the stylus. Our preliminary studies showed that using a specialized driver to control the DC motor is also critical for displaying the targeted haptic feedback. Hence, we used special purpose motor drivers for both designs (Table 1).

Table 1 – Physical components and specifications of both HaptiStylus designs

| Components | HaptiStylus 1.0 | HaptiStylus 2.0 |
|---|---|---|
| **Haptic Driver** | DRV8601, TI | DRV 2603, TI |
| **DC Motor** | Generic | Re-max 13, Maxon |
| **DC Motor Driver** | L293D, ST | DRV 8835, TI |
| **Microcontroller** | ATtiny2313-PDIP | ATtiny2313-QFN |
| **Communication** | Wire | Bluetooth 2.0 |
| **Dimensions** | Ø30mm, L=170mm | Ø18.5mm, L=192mm |

## 3.2 Generation of the Haptic Effects

The HaptiStylus can generate two haptic effects: (1) up-and-down movement effect along the stylus, (2) rotational torque effect about the long axis of the stylus.

### 3.2.1 Movement Effect

We achieve the movement effect by actuating the two vibration motors while preserving a carefully selected delay between their actuation times. Sherrick and Rogers observed that actuating two vibration motors placed in close proximity of the skin leads to one of three sensations depending on the delay between the start-up times of the actuators [1]. In particular, depending on this delay, defined as the inter-stimulus onset interval (ISOI), participants perceive either a single stationary target, two discrete stationary targets or a single moving target as shown in Figure 2. When ISOI is small, it leads to the perception of a single and stationary stimulus (Figure 2 / Single Stationary). If the ISOI value is large, it leads to the perception of two discrete stimuli near the actuators (Figure 2 / Discrete). Using a set of carefully selected ISOI values, it is possible to create the sensation of stimulus travelling from one vibration motor to the other (Figure 2 / Continuous).

The earlier studies on haptic movement effect have also focused on identifying optimal stimuli duration and ISOI values, but only on sensation through human forearm and back [12]. To our knowledge, our work is the first to investigate the perception of tactile movement on a stylus through human hand and fingers.

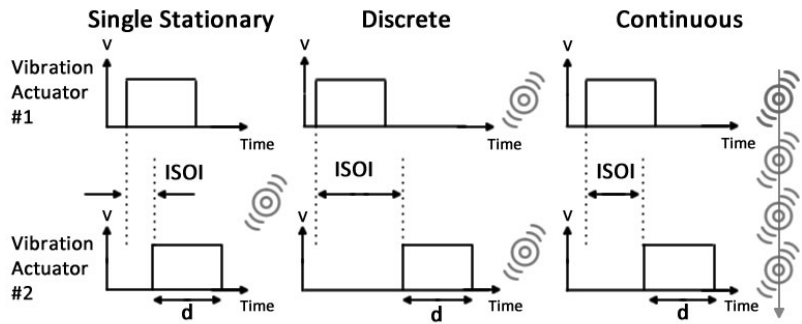

Figure 2 – Effect of ISOI for constant duration (d) on apparent tactile motion

### 3.2.2  Rotational Torque Effect

Unlike torque feedback devices, a stylus is an ungrounded apparatus, thus it is highly difficult to generate torque feedback using conventional actuators. Instead, we attempt to induce a sensation of rotation by powering the DC motor through discrete pulses. This creates a short, but perceivable sensation of rotation about the axis of the pen by slightly stretching the finger pads (Figure 3).

The phenomenon that leads to the rotational torque effect is closely related to what happens when a DC motor is powered up. When a DC motor is first powered up, the rotor experiences non-zero angular acceleration until it reaches a terminal angular velocity $\omega_{max}$ (i.e. during the on-time state). This non-zero acceleration causes a corresponding torque τ in the casing of the motor in the reverse direction, which approaches to zero as the angular velocity approaches $\omega_{max}$. When the motor is powered off, (i.e. during the off-time state), a reaction torque is generated in the opposite direction. Generating successive pulses with carefully selected frequency and duration creates a sense of rotation that is perceivable by the users. We report a comprehensive analysis of the effects of these parameters in the evaluation section.

Another parameter that affects the perception of rotation is the shape of the waveform that drives the DC motor. Based on this observation, which became evident in our preliminary experiments, we designed and conducted a controlled psychophysical experiment to compare three alternative waveforms. The results are reported in the evaluation section.

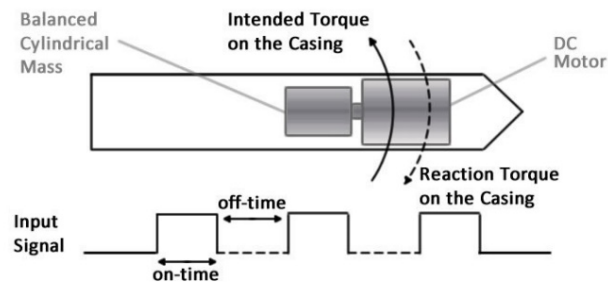

Figure 3 – During on-time, motor generates torque about intended direction and during off-time, reaction torque is generated on the casing (Balanced cylindrical mass was used for HaptiStylus 1.0 to achieve greater amount of intended torque)

## 4. EXPERIMENTAL RESULTS

For each tactile effect, we conducted separate psychophysical experiments to determine the effective values of stimulus parameters. Experimental methods and procedures for the experiments were the same. During the first two experiments, HaptiStylus 1.0 was used. Three separate groups of 10 participants took part in the experiments. The participants did not have any known sensory impairments and they were all right handed. During the experiments, the participants used their right hand to hold the stylus and left hand to enter their responses by pressing the keys on the keyboard. The participants sat comfortably on a chair facing the computer screen displaying the experimental protocol and put on headphones that played white noise to block auditory cues. In the second experiment, a visual barrier was placed to prevent the participants from seeing their hand and the stylus.

**Experiment 1: Haptic Movement Effect**

The goal of the first experiment was to determine the effective values of stimulus duration and ISOI for displaying the haptic movement effect. Based on our preliminary studies, we selected five different values, 50, 100, 200, 300, and 400 ms, for stimulus duration and also for ISOI. The experiment consisted of a total of 500 trials: 50 trials (5 stimulus durations x 5 ISOIs x 2 directions) with 10 repetitions for each trial. A total of 10 people participated in the study (5 female and 5 male, average age = 24±2). The participants completed the experiment in two sessions with a break of 3 hours between the sessions and each session took no more than 20-25 minutes. The participants were asked to characterize the tactile stimuli displayed through the vibration motors as "single stationary", "discrete", and "continuous" for each trial. To eliminate any bias during the experiment, the trials in each session were randomized. Before each session, a training session was administered, during which all possible combinations of the parameter values were presented.

**Experiment 1: Results**

Figure 4 shows the results of the first experiment. This figure was constructed based on the average responses of the participants in percentage for two directions of movement (from tip to the end of the stylus or vice versa). This concise graph shows the effect of stimulus duration and ISOI on the perception of vibrotactile stimuli. The figure also clearly illustrates the three perceived effects by the participants: the bottom left region representing the "single stationary", the top left region representing the "discrete", and the middle right region representing the "continuous" stimuli.

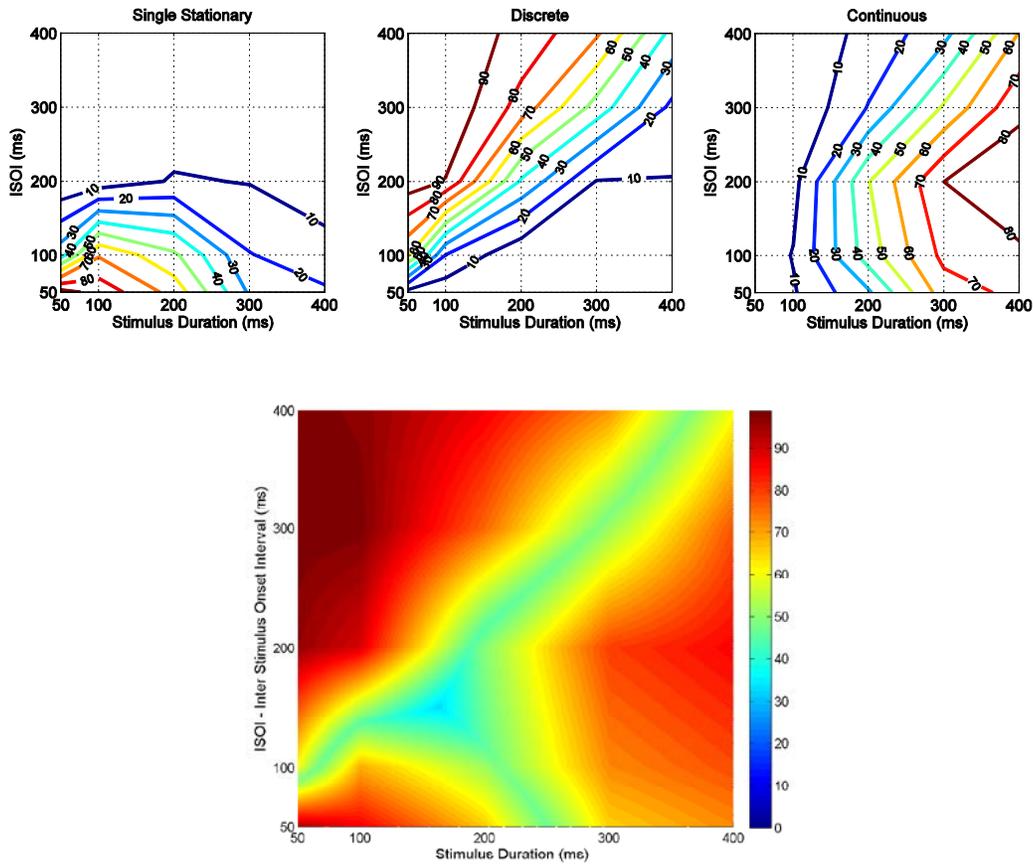

Figure 4 - Movement Effect: Mean percentage of votes for two directions of movement. The contour graphs on the top row display the percentage of the votes given by users (single stationary, discrete, and continuous) for combinations of various ISOI and stimulus duration values. The bottom graph combines these plots and shows the rating of the most dominant effect at each point in terms of percentages. Note how the valleys in the combined plot clearly divide the space into three distinct regions corresponding to distinct effects.

We investigated the effects of stimulus duration, ISOI, and the direction of tactile movement on the perception of three possible cases. The responses of the participants were analyzed by a Three-Way Repeated Measures Analysis of Variance (ANOVA) and the significance level of α=0.05 was used throughout the analysis. Our analysis shows that duration of vibration stimulus is a significant factor [$F(4,36)=102.64$, $p<0.05$] for perceiving the movement effect. Results suggested that the perception of the tactile movement effect is higher as the duration of stimulus increases ($p<0.05$). The ANOVA analysis also suggested that ISOI was a significant factor on the perception of the movement effect [$F(4,36)=7.18$, $p<0.05$]. Paired t-tests and Figure 4 suggest that the ISOI values between 50 and 200 ms were perceived as "continuous" stimuli. The direction of the tactile movement as a third parameter had no effect on the perception of the tactile movement affect [$F(1,9)=0.03$, $p=0.86$]. In conclusion, the analysis results show that stimulus duration and ISOI are effective parameters on perceiving the movement effect whereas the direction of movement has no effect.

The results of the first experiment also revealed that the effective parameter values for obtaining apparent tactile motion in a hand-held stylus differ from the values reported in the earlier studies performed on the human forearm and the back [12]. This result can be explained by the differences in the sensitivities of the human hand, the fingertips, the forearm and the back.

**Experiment 2: Rotational Torque Effect**

The goal of the second experiment was to determine effective values of on- and off-time durations of the input voltage pulses for displaying haptic rotational torque effect. Based on our preliminary studies, we selected six different values, 25, 75, 175, 275, 375, and 575 ms, for on- and off-time durations. About the long axis of the stylus, rotational torque directions of "clockwise" and "counter-clockwise" were displayed to the participants during the experiment. The experiment consisted of a total of 720 trials: 72 trials (6 on-time x 6 off-time x 2 directions) with a 10 repetitions for each trial. A total of 10 people participated in the experiment (6 female and 4 male participants, average age = 24±3). The participants completed the experiment in three sessions with a break of 2 hours between the sessions and each session took no more than 20-25 minutes. The participants were asked to differentiate the direction of rotational torque as "clockwise" or "counter-clockwise" in each trial. Before starting each session, a training session was administered exposing the participants to all possible combinations of the parameter values.

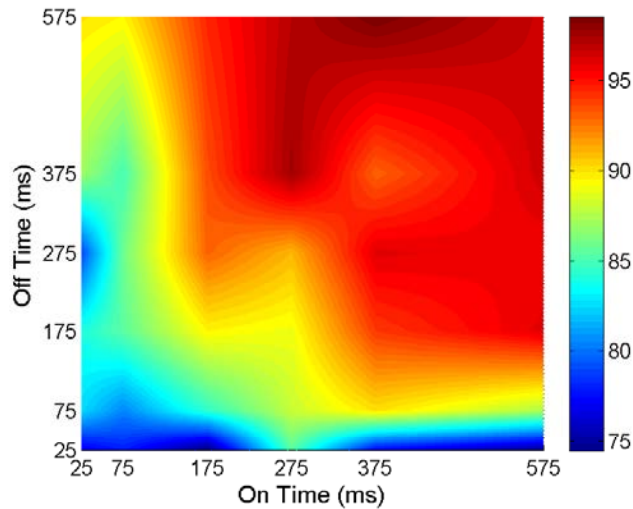

Figure 5 – Rotational Torque Effect: Mean percentage of votes for two directions of rotational torque. The tactile effect is given in a) clockwise and b) counter-clockwise directions on the stylus. The color bars next to the plots indicate the percentages of the perceived direction.

**Experiment 2: Results**

Figure 5 shows the effect of on- and off-time durations on the perception of rotational torque effect displayed in the second experiment. On the graph, the color diagram represents the percentage agreement between the intended and perceived direction of rotational torque by the participants. Three-Way Repeated Measures Analysis of Variance (ANOVA) was performed to investigate the effects of on-time duration, off-time duration and polarity of motor (used to set the intended direction of torque) on the direction perceived by the participants. A significance level of α=0.05 was used throughout the analysis. The results suggest that both on- and off-time durations significantly affect the perceived direction of rotational torque [$F(5,45)=4.49$, $p<0.05$ and $F(5,45)=13.32$, $p<0.05$, respectively]. As shown in Figure 5, the participants have identified the direction of rotational torque more successfully with higher on- and off-time durations. ANOVA analysis also suggests that participants perceived the direction of rotational torque in the intended direction of torque [$F(5,45)=0.03$, $p=0.86$].

**Experiment 3: Waveform User Study**

The haptic rotational torque effect is displayed by square shaped input pulses as described in the second experiment. We further investigate how input waveform patterns affect the perception of the rotational torque effect. Based on our preliminary studies [2], we selected three input voltage waveforms for this purpose (see Figure 6): square wave, square wave with an increasing ramp during the on-time, and square wave with a decreasing ramp during the on-time. To implement these waveforms, we used HaptiStylus 2.0, and kept the voltage applied to the DC motor, on time, and off time of the wave constant, but varied the initial and final slopes.

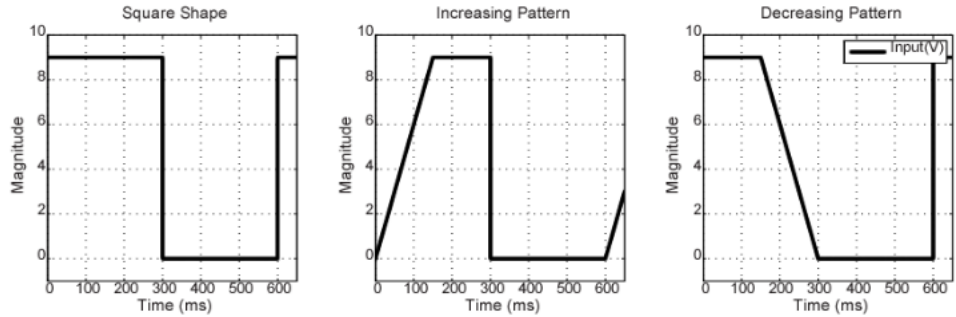

Figure 7 – "Square Shape", "Increasing" and "Decreasing" input waveform patterns for rotational torque effect.

The same experimental procedures were applied to the participants as in the second experiment. There were 10 participants (5 female and 5 male) and the average age of the participants was 25±2. We selected three different values, 50, 200, and 350 ms for on- and off-time durations based on our preliminary studies [2]. There were a total of 180 trials in this experiment: 18 trials (3 on-time durations x 3 off-time durations x 2 directions of rotational torque) with 10 repetitions of each trial. The participants completed the experiment in two sessions in the same day and each session took no more than 20 minutes. In each trial, the participants were asked to identify the torque displayed through the casing as "clockwise", or "counter-clockwise" with respect to the long axis of the stylus. The trials in each session were randomized to eliminate any bias. Before starting the experiment, the participants went through a training session exposing them to all possible combinations of the duration values and waveform patterns.

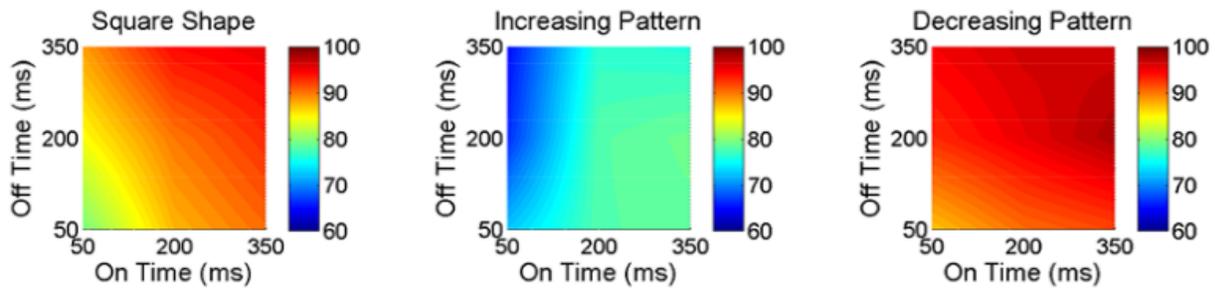

Figure 8 – Correct identification percentages for "Square Shape", "Increasing" and "Decreasing" waveform patterns.

**Results: Experiment 3**

Figure 7 shows the effect of input voltage waveform on the perception of direction of rotational torque. In each plot, red colored region represents the parameter space in which the direction of rotational torque was perceived by the participants. We also analyzed the responses of the participants by One-Way Repeated Measures Analysis of Variance (ANOVA). A significance level of $\alpha=0.05$ was used throughout the analysis. ANOVA results suggested that the waveform of the input voltage significantly affects the perception of rotational torque [$F(2, 26) = 41.96$, $p<0.05$]. The results also showed that participants have identified the direction of rotational torque more successfully when a square wave signal with a decreasing ramp during on-time was applied to the motor. The results of the third experiment clearly suggest that, in addition to the on- and off-time durations, the waveform is another significant factor on the perception of rotational torque. For example, at 200 ms on- and off-time durations, the percentages of the perceived direction are 90%, 78% and 95.5% for the three waveforms respectively (Figure 7). The increase from 90% to 95.5% can be explained by the observation that gradual offset of voltage would spread the effects of torque in the reverse direction over time and result in smaller differences in torque per unit time. We anticipate that the torque perceived in the undesired direction becomes less noticeable as the differences fall below the JND, but this topic obviously requires further investigation.

## 5. PRACTICAL USE CASES

In this section, we present two mockup demonstrations and an interactive stylus-based game to illustrate the vast potential of our flow and rotation rendering technologies in practical applications. The mockup demonstrations are supported by a supplementary video, and the discussion of the stylus-based game is supported by a rigorous user study that measures the degree to which the effects in question can be successfully rendered and perceived by the users.

### 5.1 Mockup Demonstrations

Our mockup demonstrations can be found in the supplementary video provided with the manuscript. The videos illustrate how our technology could potentially be embedded into two real-world applications to build engaging interfaces. The first application concerns a serious information visualization problem, while the second one is an interactive mobile game with rich graphical and interactive elements.

#### 5.1.1 Information Visualization

Today, we mostly use the visual channel to extract information from data. However, when the number of dimensions in a data set is large and it needs to be represented in connection with some other data sets, purely visual representations typically overwhelm the user. An excellent example for this case is the climate visualization. Climate data typically consists of values for temperature, humidity, wind intensity and direction, precipitation, and cloud water for a set of grid points in the atmosphere. Patterns in and relationships among these variables need to be examined in order to understand climate phenomena. Because the data is complex and includes a large number of dimensions, different parameters are typically shown on different maps, requiring the integration of information from multiple maps for the interpretation. This may be especially difficult for non-experts.

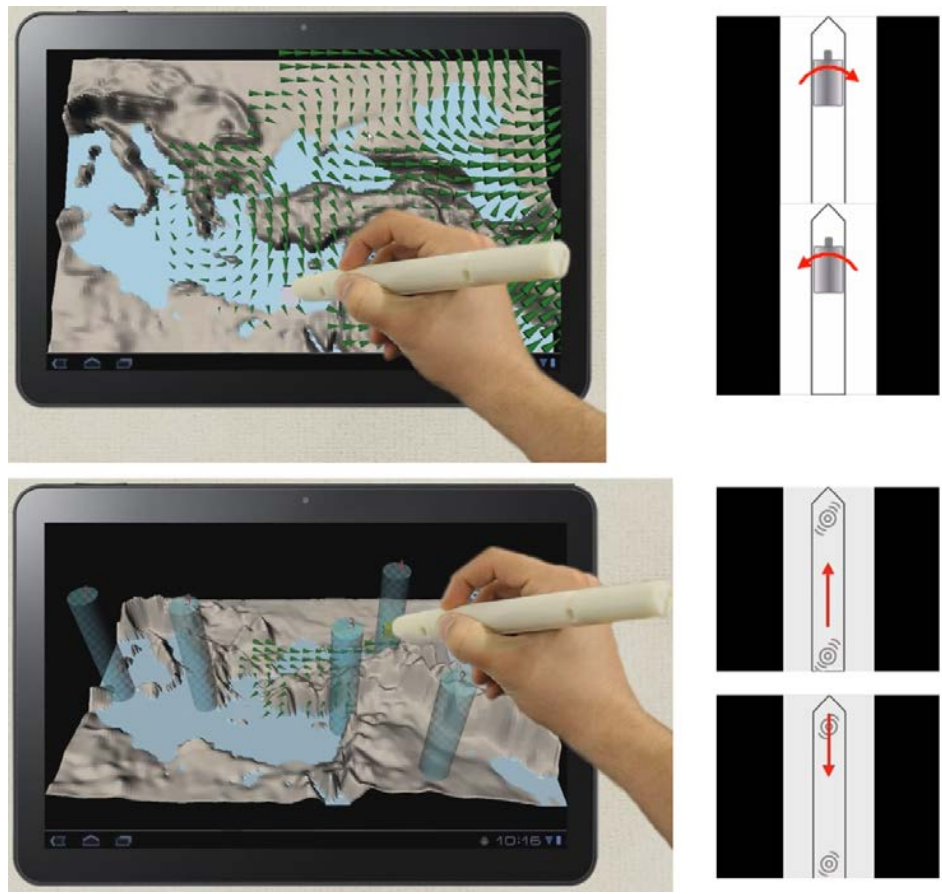

Figure 8 – Climate Visualization Mockup: Climate visualization requires conveying multidimensional information. Using the proposed stylus, direction of circulation and the rise-fall movements of the wind can be conveyed with haptic effects.

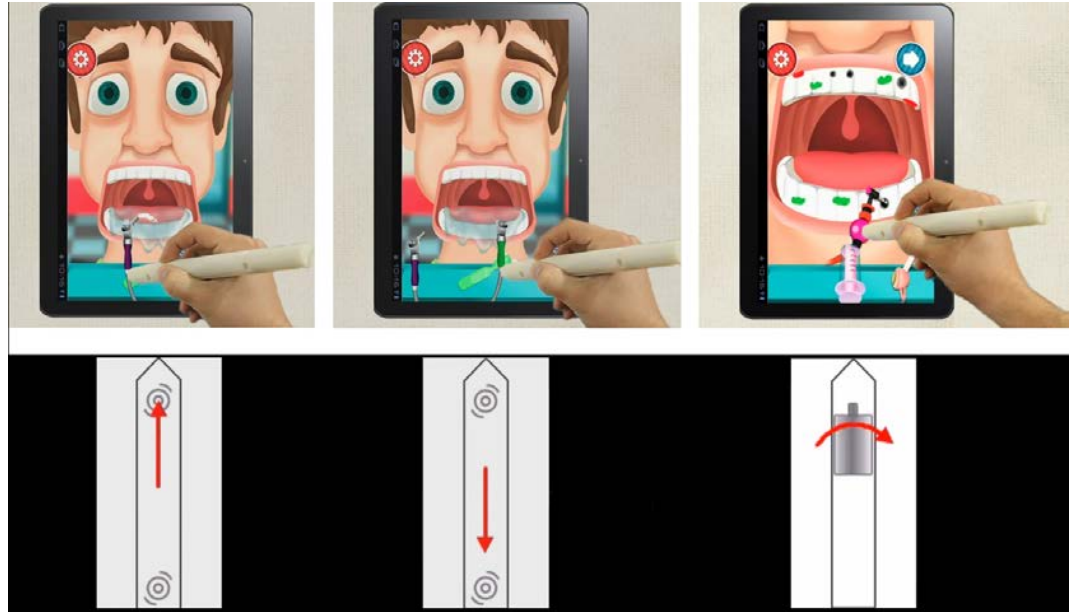

Figure 9 – Mobile Dentist Game Mockup: In this game, the player provides dental care through virtual tools. The experience of using the tools can be enhanced by haptic effects (e.g., downward flow for the water syringe, upward movement for the suction tip, and rotation for the drill).

Using the proposed stylus, some of the climate information can be displayed through the haptic channel in order to alleviate the perceptual and cognitive load of the user (see Figure 8). For example, vorticity, which represents the amount of circulation or rotation, is a force, defined as the curl of the wind velocity. The direction of this circulation can be displayed to the user via the haptic rotation effect. Moreover, the vorticity makes the wind rise or fall due to the circulation of the air, which can be displayed to the user via haptic movement effect. In this fashion, haptic feedback may improve the discovery, learning, and retention of cause-and-effect relationships in a multidimensional climate data [18].

### 5.1.2 Mobile Gaming

With the introduction of mobile devices into the market, mobile gaming has become wide-spread and increasingly more sophisticated. With the integration of rich haptic feedback to the mobile devices, these games will become more immersive, more enjoyable, and more engaging. Here, to motivate the reader, we suggest the use of haptic feedback in a mock-up dentist game designed for kids. The dentist games available in the market are virtual platforms for kids to learn oral health. They typically involve simulation of many dental procedures (cleaning, drilling, brushing, filling, suction, etc.) performed with different dental tools. They are also excellent examples demonstrating the importance of multisensory experience since it is more fun to play these games with audio is on. Hence, we believe that the haptic feedback displayed through the proposed stylus can further augment this experience by stimulating the tactile channel. For example, the tactile difference between brushing front, middle, and back teeth, tooth filling, as well as dental suction from mouth can be displayed via haptic movement effect (See Figure 9). On the other hand, tactile feeling of drilling or polishing a tooth can be displayed through the haptic rotation effect.

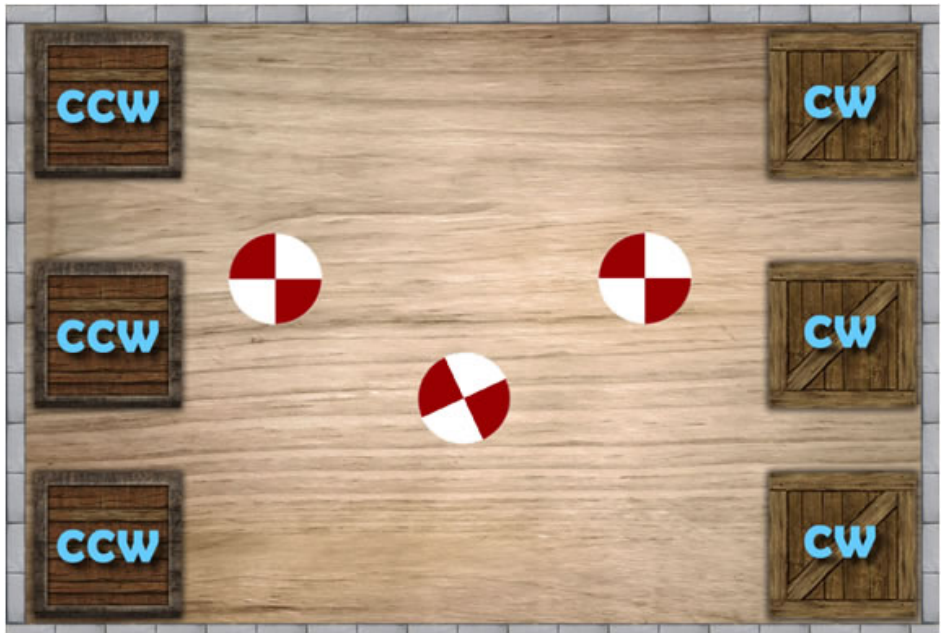
Figure 10 – Screenshot of the Spinning Tops Game.

5.2 Spinning Tops Game

To further demonstrate the practical utility of the proposed haptic effects in an application, we developed an interactive pen-based game, called "Spinning Tops", using a game engine (Game Maker, https://www.yoyogames.com/studio). A total of 15 participants played the game. The average age of the participants was 23±5. During the game, haptic movement and rotational torque effects were displayed to the participants through HaptiStylus 2.0. For this purpose, the most effective parameter values were selected based on the results of the psychophysical experiments.

As shown in Figure 10, the game scene involves 3 tops, spinning in 3 different speeds, 2 directions, and 3 boxes on each side to drop the tops. The boxes on the left and right were labeled as "CCW" and "CW", indicating the direction of spin as "counterclockwise" and "clockwise", respectively.

The goal is to identify the spinning direction of each top first, and then drag and drop it to one of the available (open) boxes on each side using visual and haptic cues. The direction of the spin (CCW versus CW) is conveyed to the user via the visual and haptic (rotational torque effect) cues while the availability of the box (open versus closed) is conveyed to the subject through the haptic movement effect only. The box is open (closed) for the drop if there is haptic movement from the distal (proximal) end of the stylus to the proximal (distal) end of the stylus.

Three sets of experiments were designed and embedded into the game:

Experiment I: We investigated the effect of haptic cues on the perception of spin direction and availability of the boxes when the visual cues in the scene are not sufficient for making decisions. For this purpose, the tops were rotated 90 degrees per frame in the visual scene and the visual scene was updated at 30 frames per second. Since the symmetrical pattern on the surface of tops repeats itself every 180 degrees, the tops appear to be stationary and the participants could not differentiate the direction of spin.

The experiment was performed under two different sensory conditions: a) NVH: no visual and haptic cues were available to the participants about the direction of spin (CCW vs. CW) and the availability of the boxes (open vs. closed). b) OH: only haptic cues are displayed to the participants about the direction of spin (via haptic rotational torque effect) and the availability of the boxes (via haptic movement effect). There were 40 trials in this experiment (2 sensory conditions x 2 spin directions x 10 repetitions). The trials were randomized and then displayed to the participants in the same order.

The results of the first experiment are given in Figure 11a and 11b. The participants make random decisions when there are no visual and haptic cues (NVH condition). The percentage of correct response is $45 \pm 13$ % (Figure 11a) and $32 \pm 10$% (Figure 11b) for the direction of spin and the availability of the box, respectively (note that the expected percentage of correct response is 50% -- one out of two directions -- for the direction of spin and 33% -one out of 3 boxes- for the availability of the box). These values increased to $81\pm19$% (Figure 11a) and $65\pm24$% (Figure 11b), respectively when the participants were provided with haptic cues (OH condition). Further analysis with One-Way Repeated Measures Analysis of Variance (ANOVA) suggests that the percentage of correct responses under NVH and OH conditions are significantly different [$F(1,14)=27.7, p<0.0001$].

Experiment II: We investigated the effect of haptic cues on the perception of spin direction when there are already sufficient visual cues in the scene to make a decision. For this purpose, the tops were rotated at 45 degrees either in the CW or CCW direction and the visual scene is updated at 30 frames per second. It was straightforward to differentiate the direction of spin by visually inspecting the scene.

The participants played the game under 2 different sensory conditions: a) OV: only visual cues were provided to the participants about the direction of spin, b) VH: Both visual and haptic (via rotational torque effect) cues were provided to the participants about the direction of spin. As shown in Figure 11c, the participants were 100% and 98% correct in their responses under OV and VH conditions, respectively and there is no significant difference between these two sensory conditions.

Experiment III: We investigated the performance of the participants when there is a mismatch between the visual and haptic cues. For this purpose, the tops were rotated at 45 and 135 degrees per frame in the CW and CCW directions, respectively. When the tops were rotated at 45 (135) degrees per frame, it was visually perceived as they were spinning in the CW (CCW) direction or vice versa. During mismatch between visual and haptic condition, the haptic cues were provided to the participants in the direction opposite to the visually perceived one, hence causing a disparity between the visual and haptic cues (MVH condition).

As shown in Figure 11d, under ambiguous sensory conditions, the participants preferred the visual cues over the haptic ones in making a decision about the direction of rotational torque. Also, One-Way Repeated Measures Analysis of Variance (ANOVA), shows that VH and MVH conditions are significantly different [F(1,14)=38.7, p<0.0001].

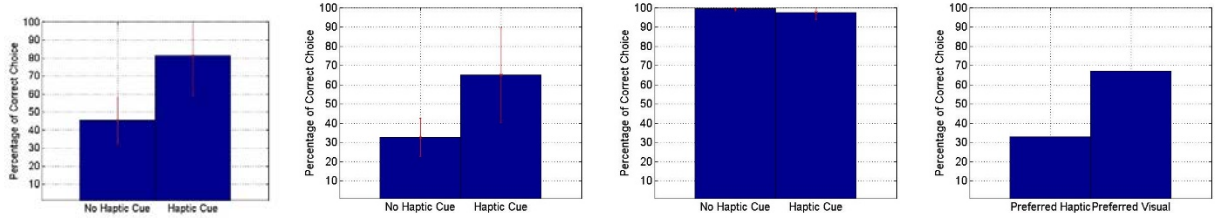

a) Rotational Torque Effect   b) Movement Effect   c) OV & VH   d) MVH

Figure 11 – Spinning Top Game: Percentage of correct choices of participants under various sensory conditions for tactile movement and rotational torque effects

## 6.  FUTURE WORK

In our experiments, we explored a comprehensive set of values for parameters that create a sense of rotation about the long axis of the stylus. However, we did not attempt to build a mathematical model of the relationship between the parameters and perception ratings. The challenge in relating input values directly to the perceived sensations lies in the fact that both the input parameters and the sensation ratings are discrete labels; however they are connected through a time varying torque signal, which is continuous. Hence, a feasible strategy might involve building a mathematical model of the relationship between input parameters and output torque, as well as the relationship between the torque profiles and the perceived effects. In order to assess the feasibility of this idea, we collected torque data for combinations of input parameter values by attaching a torque sensor on the stylus casing [2]. Preliminary results from a set of system identification experiments show that it is possible to obtain a mathematical model of the relationship between the input parameters and torque response. This leaves the task of relating the perceived user sensations to the displayed torque profiles as an interesting piece of future work. This may potentially lead to a framework for inferring optimal actuation parameters.

Another piece of future work involves improving the physical and ergonomic characteristics of the stylus. Although our second design is substantially more compact compared to our original stylus, more improvement is needed to achieve the form factor of a regular stylus. This is partly because all used parts (actuators and the batteries) were off-the-shelf items. Replacing these with custom-made versions with smaller form factors is likely to lead to an improvement. The improved form factor is likely to improve the ergonomic character of the stylus, which deserves further exploration. A possible direction might involve investigating parameters such as stylus length, center of gravity, casing texture, and pen-skin contact area.

## 7.　　CONCLUSIONS

We introduced a novel haptic stylus capable of displaying two effects to the user: the movement effect and the rotational torque effect. The movement effect was displayed by asynchronously activating two vibration motors positioned on two ends of the stylus. The rotational torque effect was displayed through a DC motor creating torque pulses in the intended direction. We explored the effective parameter spaces for both haptic effects through psychophysical experiments. The results of the psychophysical experiments using both HaptiStylus designs proved that the proposed haptic effects are perceived by the participants.

Our design opens new avenues for exploration in the pen-based computing and haptics communities. We believe that HaptiStylus can be used in a variety of mobile applications including games, entertainment, and education. For example, injection/discharge of fluids (e.g. simulating needle injection in a game), trajectory of a moving object towards/away to/from the scene (e.g. an approaching fire engine in a video scene), and penetration into a soft object (e.g. touching and feeling the softness of organs in the visual scene) can be displayed to a user using the movement effect. Similarly, opening and closing of valves and screws, sense of rotational inertia in spinning and rolling objects (e.g. flying an airplane in a game) can be conveyed to the user via the rotational torque effect. This is not an exhaustive list, but serves as evidence of the abundance of potential uses.

# REFERENCES

1. C. Sherrick and R. Rogers, "Apparent haptic movement," *Perception and Psychophysics*, vol. 1, no. 6, 1966, pp. 175-180.

2. A. Arasan, C. Basdogan, and T. Sezgin, "Haptic Stylus with Inertial and Vibro-Tactile Feedback," *Proc. World Haptics Conference* (WHC), IEEE, 2013, pp. 425-430.

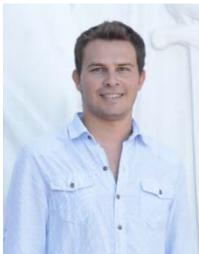

**Atakan Arasan** received his BS degree in Electrical and Electronics Engineering from Koc University in 2011. He is currently a research assistant at the Intelligent User Interfaces Laboratory and the Robotics and Mechatronics Laboratory of Koc University. His research interests include haptics, vibration and torque perception, tactile illusions, pen-based interfaces and HCI.

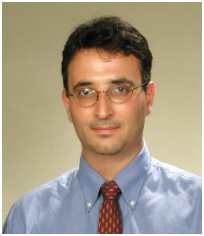

**Cagatay Basdogan** is a faculty member in the Mechanical Engineering and Computational Sciences and Engineering programs of Koc University, Istanbul, Turkey. He is also the director of the Robotics and Mechatronics Laboratory at Koc University. Before joining Koc University, he worked at NASAJPL/Caltech, MIT, and Northwestern University Research Park. His research interests include haptic interfaces, robotics, mechatronics, biomechanics, medical simulation, computer graphics, and multi-modal virtual environments. Basdogan has a PhD in Mechanical Engineering from Southern Methodist University in 1994. He is currently the associate editor of chief in IEEE Transactions on Haptics and associate editor in Computer Animation and Virtual Worlds journal.

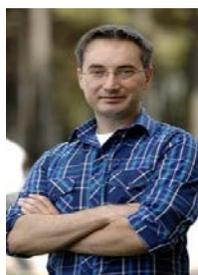

**Tevfik Metin Sezgin** graduated summa-cum-laude with Honors from Syracuse University in 1999. He received his MS and PhD degrees from Massachusetts Institute of Technology in 2001 and 2006. He subsequently joined the University of Cambridge as a Postdoctoral Research Associate, and held a visiting researcher position at Harvard University in 2010. He is currently an assistant professor at Koc University, Istanbul. His research interests include intelligent human-computer interfaces and HCI applications of machine learning. He is particularly interested in applications of these technologies in building intelligent pen-based interfaces.